\documentclass[10pt,twocolumn,letterpaper]{article}

\usepackage{iccv} 
\usepackage[accsupp]{axessibility}
\usepackage{soul}
\usepackage{soul}
\usepackage{xcolor}
\usepackage{tabularx} 
\usepackage{multirow}
\usepackage{algorithm}
\usepackage{algpseudocode}
\usepackage[export]{adjustbox}

\newcommand{\model}{\epsilon_\theta}
\newcommand{\conditioner}{\tau_\theta}
\newcommand{\expec}{\mathbb{E}}
\newcommand{\encoder}{\mathcal{E}}
\newcommand{\decoder}{\mathcal{D}}

\definecolor{iccvblue}{rgb}{0.21,0.49,0.74}
\usepackage[pagebackref,breaklinks,colorlinks,allcolors=iccvblue]{hyperref}
\usepackage{multirow}
\usepackage{soul}

\def\paperID{10723} 
\def\confName{ICCV}
\def\confYear{2025}

\definecolor{r1}{rgb}{0, 0, 0} 
\definecolor{r2}{rgb}{0, 0, 0}
\definecolor{r3}{rgb}{0, 0, 0}

\title{
Adapting Vehicle Detectors for Aerial Imagery\\
to Unseen Domains with Weak Supervision
}

\author{
    Xiao Fang\textsuperscript{1},
    \and
    Minhyek Jeon\textsuperscript{1},
    \and
    Zheyang Qin\textsuperscript{1},
    \and
    Stanislav Panev\textsuperscript{1},
    \and
    Celso de Melo\textsuperscript{2},
    \and
    Shuowen Hu\textsuperscript{2},
    \and
    Shayok Chakraborty\textsuperscript{1,3},
    \and
    Fernando De la Torre\textsuperscript{1} \and 
    \textsuperscript{1}Carnegie Mellon University, 
    \textsuperscript{2}DEVCOM Army Research Laboratory,
    \textsuperscript{3}Florida State University,\\
    {\tt\small \{xfang2, minhyekj, zheyangq, spanev\}@andrew.cmu.edu},\\
    {\tt\small \{celso.m.demelo.civ, shuowen.hu.civ\}@army.mil},
    {\tt\small shayok@cs.fsu.edu},
    {\tt\small ftorre@cs.cmu.edu}
}

\begin{document}
\maketitle

\begin{abstract}

Detecting vehicles in aerial imagery is a critical task with applications in traffic monitoring, urban planning, and defense intelligence.  Deep learning methods have provided state-of-the-art (SOTA) results for this application. However, a significant challenge arises when models trained on data from one geographic region fail to generalize effectively to other areas. Variability in factors such as environmental conditions, urban layouts, road networks, vehicle types, and image acquisition parameters (e.g., resolution, lighting, and angle) leads to domain shifts that degrade model performance. 
This paper proposes a novel method that uses generative AI to synthesize high-quality aerial images and their labels, improving detector training through data augmentation. Our key contribution is the development of a multi-stage, multi-modal knowledge transfer framework utilizing fine-tuned latent diffusion models (LDMs) to mitigate the distribution gap between the source and target environments.  
Extensive experiments across diverse aerial imagery domains show consistent performance improvements in $\text{AP}_{50}$ over supervised learning on source domain data, weakly supervised adaptation methods, unsupervised domain adaptation methods, and open-set object detectors by 4-23\%, 6-10\%, 7-40\%, and more than 50\%, respectively. Furthermore, we introduce two newly annotated aerial datasets from New Zealand and Utah to support further research in this field.
Project page is available at: 
\url{https://humansensinglab.github.io/AGenDA}

\end{abstract}    
\begin{figure} 
    \centering  
    \includegraphics[width=\columnwidth]{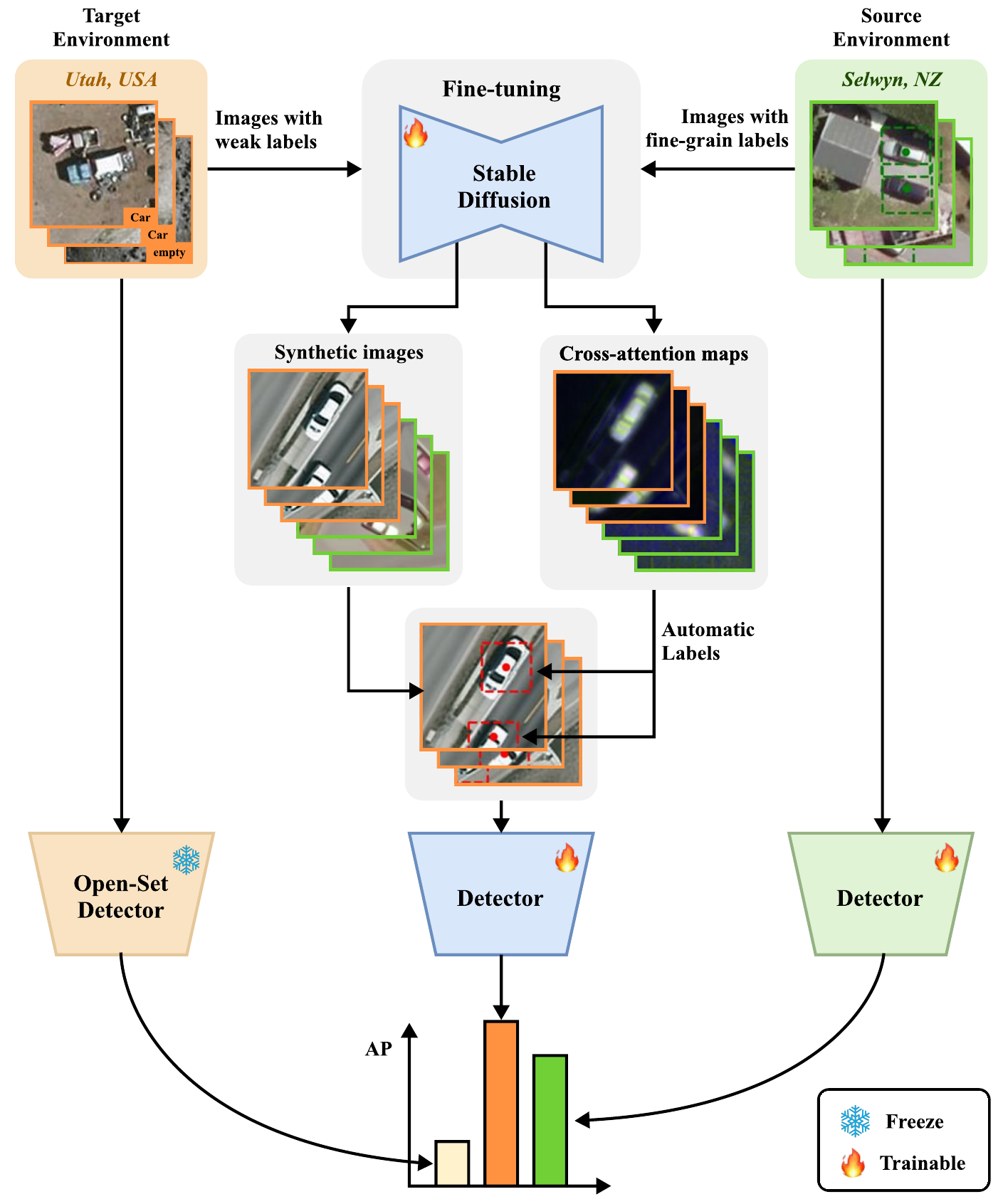}
    \caption{
        \textbf{Overview.} We propose a pipeline that generates high-quality aerial images along with their labels. Our method outperforms baseline detectors trained on source images and open-set detectors directly inferred on target images.
    }
    \label{fig:teaser}
\end{figure}

\section{Introduction}
\label{sec:intro}

Recent advancements in Generative AI have led to the development of diffusion-based models \cite{DiffusionModels} that can synthesize images with unprecedented realism, making them nearly indistinguishable from real-world data \cite{DDPM, DDIM, ImprovedDDPM}.
Additionally, integrating natural human language has become crucial to modern vision systems. This integration can take the form of conditioning the generation process (for example, in text-to-image generative models \cite{rombach2022high, podell2024sdxl} utilizing cross-attention mechanism) or serving as an additional modality that shares a common latent space with vision (as seen in models like \textit{CLIP}~\cite{CLIP}).
One key reason for the success of these new models is that they have been trained on extensive datasets, such as \textit{LAION-5B}~\cite{schuhmann2022laionb} or \textit{WIT}~\cite{CLIP}, consisting of hundreds of millions or billions of text-image pairs. As a result, the scientific community often refers to them as \textit{Foundational Models}.


Upon their inception, researchers have explored how to use foundational models to generate training data for downstream tasks like classification, object detection, and semantic segmentation. Methods such as \textit{ALIA}~\cite{ALIA},  \textit{DatasetDM}~\cite{wu2023datasetdm}, \textit{Dataset Diffusion}~\cite{nguyen2023dataset}, \textit{DiffusionEngine}~\cite{zhang2023diffusionengine} are particularly useful in scenarios with limited or no access to natural training data.

Despite the promising results these methods have achieved augmenting general-purpose datasets like \textit{VOC}~\cite{pascal-voc-2012} and \textit{COCO}~\cite{COCO}, little to no effort has been made to generate diverse, high-resolution annotated synthetic datasets on aerial views. These datasets are important for training detectors on small objects like vehicles.
The primary reason for this gap is that off-the-shelf generative models,
struggle to generate these views due to insufficient aerial imagery representation in their training datasets. Hence, fine-tuning is unavoidable in this case. 
Recent efforts have primarily produced generative models constrained by geospatial resolution, such as \textit{DiffusionSat}~\cite{khanna2024diffusionsat} and \textit{SatDiffMoE}~\cite{song2024satdiffmoe}, and therefore cannot be utilized for vehicle detection.

\textcolor{r3}{The lack of sufficient large-scale aerial view training data impacts the performance of not only generative foundational models but also other types of \textit{Vision Large Language Models~\cite{mllmsurvey}} (VLLMs), such as 
\textit{BLIP2}~\cite{li2023blip2}, \textit{InternVL3}~\cite{internvl3}, and \textit{Gemini}~\cite{team2023gemini}. Our analysis reveals that these models perform poorly in zero-shot settings for tasks such as car presence classification and center localization in aerial imagery. (see Appendix~\cref{sec:supp-failure_case}). This limitation also extends to open-set object detectors~\cite{li2021grounded, zhao2024realtimetransformerbasedopenvocabularydetection, minderer2024scalingopenvocabularyobjectdetection}, which, as shown in \Cref{tab:cross-domain-sota-satellite}, exhibit unsatisfactory results on aerial imagery.}

On the other hand, detection
annotations should also be produced for each generated image object. 
However, text-to-image models do not naturally provide these.
There are two potential options to address this challenge:
1) the annotations (\eg, semantic segmentation maps, bounding box layouts) to be created before the image generation and used as a spatial conditioning signal~\cite{ControlNet, GLIGEN}, 
or 
2) products from the generative process, such as cross- and self-attention maps extracted from the denoising U-Net~\cite{DDPM} to be employed to image produce annotations~\cite{tang-etal-2023-daam, wu2023datasetdm, nguyen2023dataset, zhang2023diffusionengine}. 
In the first case, studies reported that the generative models often do not fully obey the imposed conditioning, especially for small objects such as vehicles in aerial view images \cite{SOEDiff}, which may result in generated annotations that do not match the corresponding synthetic images. Therefore, we studied the prospects of the second approach.

To this end, we introduce a new method for annotated aerial view vehicle detection via synthetic image dataset generation, which employs stacked (multi-channel) cross-attention maps, learnable text prompt tokens, and multi-stage cross-environment (source to target) knowledge transfer. We consider the scenario where a fully annotated dataset with bounding box annotations from the source environment and a target environment dataset with weak binary annotations (whether or not a vehicle is present in a given image) are available, which are much easier to obtain than full bounding box annotations. 
Our extensive experiments show considerable improvements over baseline (model trained only on the source environment data) performance demonstrated by four popular object detectors.

Our contributions can be summarized as follows: 
\begin{itemize}
    \item We introduce a novel approach for generating aerial view annotated synthetic image datasets based on multi-channel cross-attention maps and multi-stage cross-environment knowledge transfer.
    \item We conduct extensive experiments involving four popular state-of-the-art object detectors to assess the effectiveness of our approach. We compared the performance of our method with other published methods for open-set detection, unsupervised object detector adaptation, and weakly supervised model adaptation. The results corroborate the promise and potential of our framework.
    \item We introduce two new real-world aerial view datasets captured in Selwyn (New Zealand), 2,078,077 images, and Utah (USA), 2,684,658 images, including car (small vehicle) detection task annotations.  
\end{itemize}

\section{Related Work}
\label{sec:related_work}

\subsection{Diffusion Models for Perception Tasks}

Diffusion models~\cite{DDPM,dhariwal2021diffusion} have undergone significant developments and have emerged as prominent generative models in modern research. These models work by progressively corrupting data with noise, and then learning to reverse this process to reconstruct the original data. Once trained, new data can be generated by applying the learned denoising process to randomly sampled noise.
Latent diffusion models (LDM)~\cite{rombach2022high,podell2024sdxl} perform diffusion process in the latent image space, which reduces the computation cost towards high-resolution image synthesis. 
Customized image generation can be achieved by adding various types of control,  such as texts~\cite{ruiz2023dreambooth}, edges~\cite{ControlNet}, segmentation masks~\cite{park2024shape}, geometric
layouts~\cite{chen2024geodiffusion},  and images~\cite{ramesh2022hierarchicaltextconditionalimagegeneration}.

Recent research demonstrates that diffusion models can also be utilized in various perception tasks. Diffusion Classifier~\cite{li2023your} reveals that pre-trained diffusion models can be directly employed for zero-shot classification tasks. DoGE~\cite{wang2024domain} conditions Stable Diffusion~\cite{rombach2022high} on CLIP image embedding difference between two domains, improving classification and semantic segmentation accuracy. DGInStyle~\cite{jia2024dginstyle} combines semantic
masks with style prompts to generate training data for semantic segmentation. DatasetDM~\cite{wu2023datasetdm} uses a few labeled real images to train
a mask decoder, leading to a robust synthetic data generator. Thanks to the cross-attention mechanism in Stable Diffusion, several works~\cite{wu2023diffumask,wang2024diffusionmodelsecretlytrainingfree, nguyen2023dataset} obtain high-quality segmentation labels through cross-attention maps between images and target concepts. AttnDreambooth~\cite{pang2024attndreambooth} further incorporates learnable tokens before target concepts to enhance the accuracy of cross-attention maps.
Compared to previous works, our method differs in
several aspects. First, we explore Stable Diffusion for cross-domain object detection where diffusion models are rarely trained, while previous works mostly focus on classification and segmentation. Second, we synthesize accurate labels by introducing learnable tokens to encode foreground and background concepts for more accurate cross-attention maps and labeling based on style-less cross-attention maps to remove domain gaps.

\subsection{Cross-domain Object Detection}
Cross-domain object detection addresses the challenge of detecting objects when a domain shift exists between the source and target environments. It can be categorized into subfields based on the availability of data and labels during the training and testing phases. 

\textit{Open-set detection} is helpful in detecting novel categories in the target domain since it identifies both known and unknown objects during inference. One approach is aligning textual and visual features by jointly training object detection ~\cite{li2021grounded}. Another approach includes integrating transformer-based architectures or YOLO framework with open-set object detection, improving robustness in detecting objects without category constraints~\cite{liu2024groundingdinomarryingdino, Cheng2024YOLOWorld}. Vision transformer-based models ~\cite{minderer2022simpleopenvocabularyobjectdetection, minderer2024scalingopenvocabularyobjectdetection} further enhance open-vocabulary object detection by utilizing large-scale text-image pre-training, enabling models to detect beyond predefined categories.
More recent research ~\cite{du2024lami, zhao2024realtimetransformerbasedopenvocabularydetection} refines open-set detection through improved representation learning and domain generalization techniques. 

\textit{Unsupervised domain adaptation} for object detection aims to improve detection performance in an unlabeled target domain by leveraging various techniques. A common approach is style transfer~\cite{parmar2024onestepimagetranslationtexttoimage, xu2023cyclenet, wang2024domain}, which aligns the visual characteristics of source and target domains to mitigate domain shifts. Recent research also uses knowledge distillation~\cite{aldi, zhou2022ssdayolosemisuperviseddomainadaptive}, where knowledge from a well-trained teacher model is transferred to a student model, enabling adaptation to the target domain. Other methods include adversarial feature learning~\cite{TIA, li2022cross} which minimizes domain discrepancies through adversarial training, or graph reasoning~\cite{li2022sigma} that captures structural relationships between objects, enhancing robustness in cross-domain object detection.

\textit{Weakly supervised domain adaptation} enhances object detection in target domains where only limited supervision is available and can be categorized into network-based methods and generative models. Network-based methods improve feature alignment and domain generalization through approaches such as hierarchical feature learning~\cite{Xu_2022_CVPR}, transformer-based adaptations~\cite{wsdetr}, and pseudo-labeling ~\cite{inoue2018crossdomainweaklysupervisedobjectdetection}. These methods refine object detection models by utilizing structured architectural modifications that help bridge domain gaps. Generative models~\cite{kondapaneni2024tadp, jia2024dginstyle} enhance adaptation by learning domain-invariant representations through style transfer and task-adaptive pre-training. By synthesizing target-like visual features, these approaches help models generalize more effectively to new environments with limited supervision, mitigating domain discrepancies while preserving task-relevant information.

\begin{figure*}[t]
    \centering  \includegraphics[width=\textwidth]{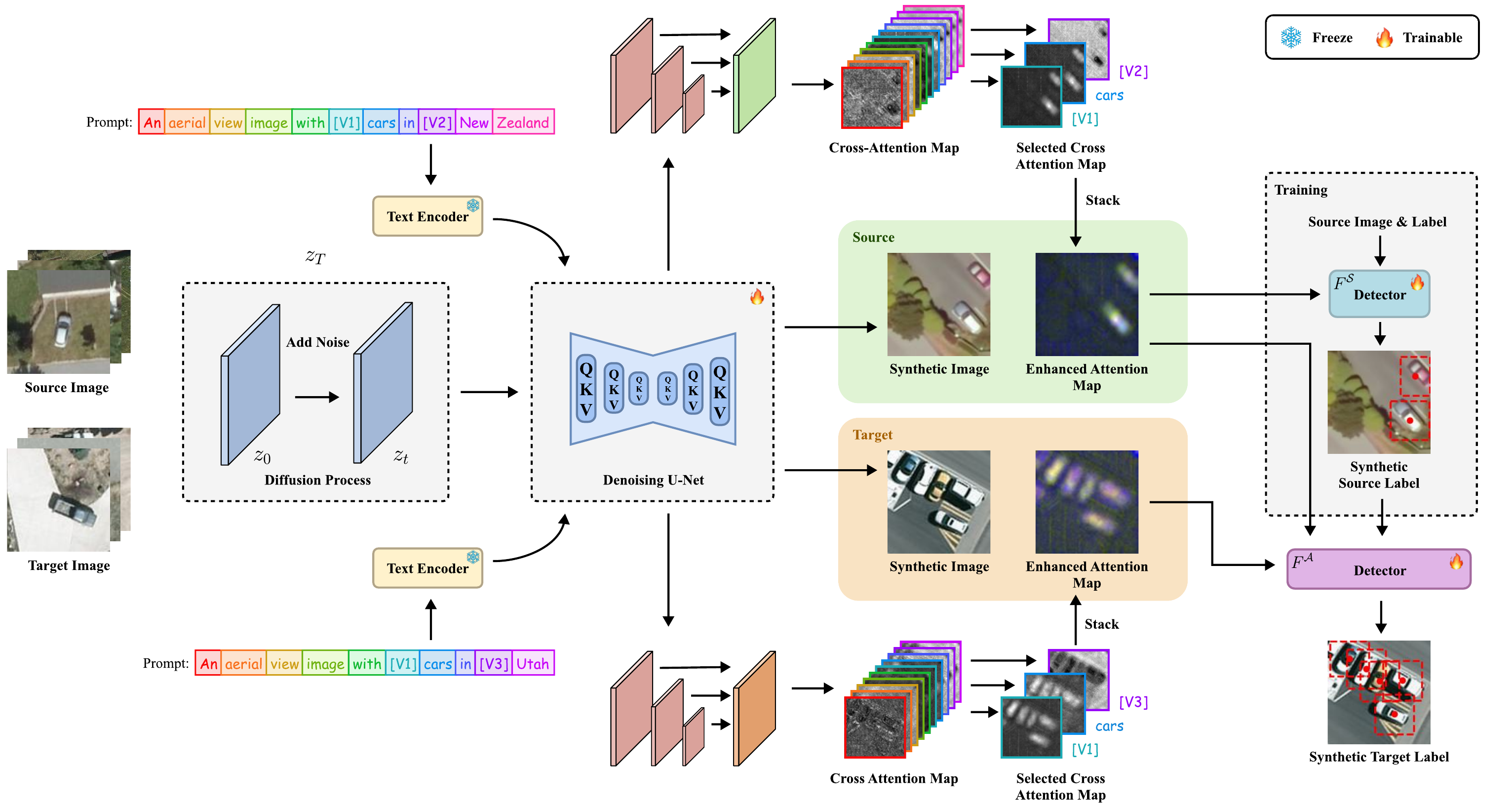}
    \caption{\textbf{Overview of our pipeline.} It consists of two stages. First, we finetune Stable Diffusion and synthesize both source and target domain images. Second, we automatically label synthetic target domain images via cross-attention maps. }
\label{fig:pipeline}
\end{figure*}

\section{Method}
\label{sec:method}

Given the fully annotated (where all vehicles in each image are annotated with bounding boxes) source domain data represented as $\mathbf{D}^\mathcal{S}=\left\{\left(x_i^\mathcal{S},y_i^\mathcal{S}\right)\right\}_{i=1}^{N_\mathcal{S}}$, and image-level annotated (whether or not a vehicle is present in a given image) target domain data $\mathbf{D}^\mathcal{T}=\left\{\left(x_i^\mathcal{T},y_i^\mathcal{T}\right)\right\}_{i=1}^{N_\mathcal{T}}$, the goal is to  synthesize a fully annotated dataset $\mathbf{D}^\mathcal{G} = \left\{\left(x_i^\mathcal{G},y_i^\mathcal{G}\right)\right\}_{i=1}^{N_\mathcal{G}}$ in the target domain, where
$x$ and $y$ denote the images and their corresponding labels, respectively. $N_\mathcal{S}$, $N_\mathcal{T}$ and $N_\mathcal{G}$ denote the total number of images in each dataset.
In \cref{subsec:text-to-image generation}, we present our approach for generating synthetic images with Stable Diffusion~\cite{rombach2022high} by fine-tuning the model to adapt to both source and target domains. In \cref{subsec:concept localization}, we leverage multi-channel cross-attention maps for object localization in synthetic images. 
In \cref{subsec:automatic labeling}, we propose an automatic labeling strategy to generate pseudo labels for synthetic target domain data. \textcolor{r3}{A summary of the full pipeline is provided in the Appendix \cref{sec:supp-method}.}

\subsection{Text-to-image generation}
\label{subsec:text-to-image generation}
Stable Diffusion consists of an image encoder $\encoder$, a conditional U-Net $\model$, a text encoder $\conditioner$, and a latent decoder $\decoder$.
We fine-tune the U-Net $\model$ on both $\mathbf{D}^\mathcal{S}$ and  $\mathbf{D}^\mathcal{T}$.  In the forward process, an image $x$  is encoded
into a latent representation $z_0 = \encoder{(x)}$. A noisy latent $z_t$ at any time $t$ is then sampled using the following sampling function~\cite{DDPM}:
\begin{equation}
    z_t =  \sqrt{\Bar{\alpha}_t} z_0 + \sqrt{1-\Bar{\alpha}_t} \varepsilon,\ \varepsilon \sim \mathcal{N}(0, \mathbf{I}), 
\end{equation} where   $\Bar{\alpha}_t = \prod_{s=1}^{t} \alpha_s$, and $t$ is uniformly sampled from $\{1, . . . , T\}$. 
To learn the reverse process, the latent $z_t$ is passed to the U-Net $\model$, along with the timestep $t$ and the
prompt embedding $\conditioner(c)$. To encode domain-specific information, we design distinct prompts for source and target domain images, denoted as $c_\mathcal{S}$ and $c_\mathcal{T}$. $c_\mathcal{S}$ and $c_\mathcal{T}$ follow the format of  ``an aerial image with [category] in [S]'' and ``an aerial image with [category] in [T]'', where [category] represents the object type, and [S] and [T] are unique identifiers to distinguish between the source and target domain prompts, as inspired by~\cite{ruiz2023dreambooth}. The U-Net $\model$ is trained to predict the noise added to the latent $z_0$: 
\begin{equation}
    L_\text{LDM} = \expec_{\encoder(x), c, \epsilon, t }\Big[ \Vert \epsilon - \model(z_{t},t, \conditioner(c)) \Vert_{2}^{2}\Big] \
\end{equation}
where $c \in \{c_\mathcal{S}, c_\mathcal{T}\}$ depends on the domain of the input image $x$.
During image generation, a pure noise latent $z_T$ is iteratively denoised through the U-Net $\model$ for $T$ steps, followed by decoding via the latent decoder $\decoder$ to
generate the final image. By conditioning the U-Net $\model$ on distinct prompts $c_\mathcal{S}$ and $c_\mathcal{T}$, we can independently synthesize images corresponding to the source and target domains.

\subsection{Concept Localization}
\label{subsec:concept localization}
It has been observed in~\cite{tang-etal-2023-daam} that in a well-trained text-to-image diffusion model, cross-attention maps between the feature maps and the conditioning text assign higher weights to regions that align with the text concept. This indicates that cross-attention maps effectively help locate the concept. Therefore, we propose
leveraging cross-attention maps corresponding to ``[category]" in a fine-tuned Stable Diffusion to facilitate
downstream visual tasks. 
Specifically, 
cross-attention maps can be obtained in each of the four layers in the U-Net $\model$, which corresponds to four different resolutions. We compute the attention map $\mathcal{A}$ by averaging all the cross-attention maps across these resolutions. Finally, we normalize $\mathcal{A}$ to $(0,1)$ to emphasize regions with higher attention weights.

While cross-attention maps for the object category effectively highlight relevant regions, their robustness can be further improved by cross-verifying with additional maps. Inspired by~\cite{gal2023an, pang2024attndreambooth},
we introduce two learnable tokens in the prompt to generate complementary cross-attention maps. These tokens are designed to capture both the object and the background concept, which includes all regions outside the objects. The insight is that combining a learnable context token with the object category enhances the localization of the target objects, while a background token helps identify non-target regions. By effectively localizing the background, we can further refine object delineation. To implement this approach, we design a two-stage pipeline. Throughout the two stages, we set the prompt as ``an aerial view image with [$V_1$] [category] in [$V_2$] [S]'' for source domain data and ``an aerial view image with [$V_1$] [category] in [$V_3$] [T]'' for target domain data. [$V_1$] represents the learnable token for the object concept, while [$V_2$] and [$V_3$] correspond to the learnable token for the source domain and target domain background concept, respectively.

In the first stage, we fine-tune both the U-Net $\model$ and learnable tokens [$V_1$], [$V_2$], and [$V_3$] to capture the new concepts. To facilitate this process, we introduce a novel cross-attention map regularization loss that encourages similarity between
the attention maps of $[V_1]$ and [category], while penalizing similarity between
the attention maps of $[V_2]$, $[V_3]$ and [category]. We denote the normalized cross-attention map of [$V_1$], $[V_2]$, [$V_3$], and [category] as $A_{\text{V}_1}$, $A_{\text{V}_2}$, $A_{\text{V}_3}$ and $A_\text{c}$, respectively. To enforce similarity between $A_{\text{V}_1}$ and 
$A_\text{c}$, we further normalize them into discrete probability distributions where pixel values sum to one while preserving their relative differences. We denote them as $\hat{A}_{\text{V}_1}$ and $\hat{A}_\text{c}$.
We then employ the total variation distance metric to minimize the difference between $\hat{A}_{\text{V}_1}$ and $\hat{A}_\text{c}$, which is equivalent to half of the $L_1$ distance between them~\cite{levin2017markov}:
\begin{equation}
    L_{\text{obj}} = \frac{1}{2}{\sum_{x,y}|\hat{A}_{\text{V}_1}(x,y) - \hat{A}_\text{c} (x,y)|}
\end{equation}
Similarly, to penalize similarity between $A_\text{bg}$ and $A_\text{c}$ where $A_\text{bg} \in \{A_{\text{V}_2}, A_{\text{V}_3}\}$, which is equivalent to enforce similarity between $A_\text{bg}$ and $A^*_\text{c}$ where 
$A^*_\text{c} = 1 - A_\text{c}$, we first normalize both $A_\text{bg}$ and $A^*_\text{c}$ into discrete probability distributions and then compute the distribution difference using the total variation difference metric as follows:
\begin{align}
    &L_{\text{bg}} = \frac{1}{2}{\sum_{x,y}|\hat{A}_\text{bg}(x,y) - \hat{A}^*_\text{c} (x,y)|}
\end{align}
The total loss function can be formulated as $L = L_{\text{LDM}} + L_{\text{obj}}+L_{\text{bg}}$.

In the second stage, we fix the learned tokens [$V_1$], [$V_2$], and [$V_3$], and further fine-tune the U-Net $\model$. This is because Stable Diffusion learns to fit the data distribution with varying prompts in the previous stage, which might not align well with the data distribution conditioned on the final learned prompt. We continue to apply the loss function $L$ to guide the learning of the attention maps and avoid any embedding misalignment.

\subsection{Automatic Labeling via Cross-Attention Maps}
\label{subsec:automatic labeling}

After fine-tuning Stable Diffusion as described in ~\cref{subsec:text-to-image generation} and ~\cref{subsec:concept localization}, we generate synthetic source domain data $\mathbf{D}^\mathcal{GS}=\left\{\left(x_i^\mathcal{GS},\widetilde{A}_i^\mathcal{GS}\right)\right\}_{i=1}^{N_\mathcal{GS}}$ and target domain data $\mathbf{D}^\mathcal{GT}=\left\{\left(x_i^\mathcal{GT},\widetilde{A}_i^\mathcal{GT}\right)\right\}_{i=1}^{N_\mathcal{GT}}$. For each synthetic data sample $D_i \in \mathbf{D}^\mathcal{GS} \cup \mathbf{D}^\mathcal{GT}$, we extract the cross-attention maps $A_{i,\text{c}}$, $A_{i,{\text{V}_1}}$ and $A_{i,\text{bg}}$ during denoising steps. The enhanced cross-attention map $\widetilde{A}_i$ is then obtained by stacking these components. Since cross-attention maps are grayscale images that highlight object regions, they contain less style information than RGB images. Therefore, we propose using these maps to generate bounding box annotations for target domain data. First, we train a detector ${F}^\mathcal{S}(\cdot; \theta)$ on the fully annotated real source domain data $\mathbf{D}^\mathcal{S}$. This detector is then used to predict reliable pseudo labels $\left\{y_i^\mathcal{GS}\right\}_{i=1}^{N_\mathcal{GS}}$ on the synthetic source domain images $\left\{x_i^\mathcal{GS}\right\}_{i=1}^{N_\mathcal{GS}}$. Next, we train another detector ${F}^\mathcal{A}(\cdot; \theta)$ on the combined dataset $\mathbf{D}_*^\mathcal{GS}=\left\{\left(\widetilde{A}_i^\mathcal{GS}, y_i^\mathcal{GS}\right)\right\}_{i=1}^{N_\mathcal{GS}}$, which comprises synthetic source domain enhanced cross-attention maps and their corresponding pseudo labels. Finally, we use the well-trained detector ${F}^\mathcal{A}(\cdot; \theta)$ to test the target domain cross-attention maps $\left\{\widetilde{A}_i^\mathcal{GT}\right\}_{i=1}^{N_\mathcal{GT}}$ and predict a set of pseudo labels $\left\{y_i^\mathcal{GT}\right\}_{i=1}^{N_\mathcal{GT}}$ for synthetic target domain data $\mathbf{D}^\mathcal{GT}$. This results in a fully annotated synthetic target domain dataset $\left\{\left(x_i^\mathcal{GT},y_i^\mathcal{GT}\right)\right\}_{i=1}^{N_\mathcal{GT}}$, which can be directly used for downstream object detection tasks.


Determining the confidence score threshold for bounding box labels is challenging because it can vary across datasets. To address this, we propose a classifier refinement method that selects more reliable labels for synthetic target domain data, inspired by~\cite{Tang_2017_CVPR}. Predicted bounding boxes with high confidence scores represent foreground objects, while those with low scores represent background regions. We define bounding boxes above a high threshold $\lambda_{\text{high}}$ as positive samples and those below a low threshold $\lambda_{\text{low}}$ as negative samples. We train a classifier using these samples on cropped image patches, refining predictions for samples with intermediate confidence scores, resulting in more reliable labels for synthetic target domain data.
\section{Datasets}
\label{sec:datasets}

\begin{figure}[!tbp]
    \centering  \includegraphics[width=\columnwidth]{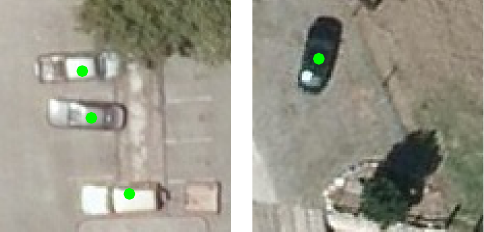}
    \caption{\textbf{Image samples from our datasets.} (\textit{left}) LINZ sample, (\textit{right}) UGRC, (\textit{green markers}) small vehicle location annotations. For more examples, check the Supplementary Material.}
    \label{fig:linz-utah-examples}
\end{figure}

We use three real-world aerial view vehicle detection datasets for our experiments---the publicly available \textit{DOTA}~\cite{Xia_2018_CVPR}, and two additional datasets we created, 
called \textit{LINZ} and \textit{UGRC} (\Cref{fig:linz-utah-examples}). All three have \textit{ground sampling distance} (GSD) of 12.5 cm per px and have been sampled to 112~px $\times$ 112~px image size.
Utilizing this image size is essential for our method because diffusion models are known for struggling with generating small objects caused by the cross-attention mechanism's limited resolution. Thus, we increase the relative object size within the images.
We use only one object class (\textit{small vehicle}) and location labels (object centers).

\noindent\textbf{DOTA:}
We used the \textit{DOTA-v2} dataset's 
training and validation sets, which contain 1,830 and 593 images, respectively,  with 169,268 and 56,062 unique small car instances.
The image sizes range from 421~px $\times$ 346~px to 29,200~px $\times$ 27,620~px.
We only used images with GSD of 15 cm per px or less and scale them to match the target 12.5 cm per px. 
The images with larger GSD or whose metadata do not provide such information are rejected.
Finally, we produce our experiments' training and validation sets by placing randomly rotated square sampling windows with size 112~px $\times$ 112~px uniformly distributed across each original image area. This results in 455,099 training images (28,453 contain objects) and 142,262 validation images (9,460 contain objects), with 62,178 and 20,194 total object instances, respectively. The original bounding box labels were converted to object locations to match the annotations of the other two datasets described below.

\noindent\textbf{LINZ:}
We created this dataset by manually annotating vehicle locations in aerial images captured in \textit{Selwyn, New Zealand} in December 2012. They were obtained from the \textit{Land Information New Zealand} (LINZ) online platform~\cite{LINZ-Selwyn-0.125m-2012-2013}. 
The original tiles' size is 240~m $\times$ 360~m (1920~px $\times$ 2880~px) with GSD of 12.5~cm per px. A uniformly distributed sampling window produces the final samples with size 112~px $\times$ 112~px.
We finally have 1,451,144 training images (19,564 of which include objects), 188,744 validation images (2,108 of which include objects), and 438,189 test images (2,629 of which include objects). The number of object instances in the training set is 29,495, in the validation set, 2,574, and in the test set, 3,640. The Appendix~\cref{sec:supp-dataset} provides more details.

\noindent\textbf{UGRC:}
We used the same approach as with \textit{LINZ} to construct this dataset. We obtained 40 high-resolution (12.5 cm per px) aerial image tiles captured in \textit{Utah, USA} and annotated the locations of small vehicle instances. The images were downloaded from \textit{Utah Geospatial Resource Center} (UGRC)~\cite{UGRC-HRO-2012} online platform.
Their native size is 2,000~m $\times$ 2,000~m (16,000~px $\times$ 16,000~px). After performing dataset sampling using a 112 px square window, we constructed splits with the following quantities: 2,142,849 training images (15,631 of which include objects), 271,252 validation images (3,912 of which include objects), and 270,557 test images (1,510 of which include objects). The number of object instances in the training set is 26,001, in the validation set 9,878, and in the test set 1,900. More information can be found in the Appendix~\cref{sec:supp-dataset}.

\section{Experiments}
\label{sec:experiments}



\subsection{Implementation Details}


\noindent\textbf{Working with location-based annotations:}
As discussed in \cref{sec:datasets}, 
LINZ and UGRC datasets contain small vehicle location labels. DOTA's bounding box annotations were converted to location labels for consistency.
To leverage the abundance of bounding box-based open-source object detection frameworks, we reformulate the location detection problem as a bounding box detection task. Specifically, we define a decision circle with a 12~px radius centered on the vehicle’s centroid.
A predicted center is considered a true positive if it falls within this circle. Then we introduce a 42.36~px square pseudo-bounding box centered at the vehicle’s centroid.
This specific size is chosen to facilitate the evaluation using the $\text{AP}_{50}$ object detection metric.
Mathematically, this approach is functionally equivalent to using the aforementioned decision circle, with minimal error. 

\noindent\textbf{Automatic Labeling:}
To establish detection baselines, we employ the MMDetection framework~\cite{mmdetection},
and evaluate a diverse set of models, including  Faster-RCNN~\cite{faster-rcnn}, 
 YOLOv5~\cite{yolov5}, YOLOv8~\cite{reis2023real} and ViTDet~\cite{vitdet},
 which represent one-stage, two-stage and transformer-based object detectors. 
 The automatic labeling process follows the methodology outlined in section~\ref{subsec:automatic labeling}, ensuring consistency by utilizing the same detector throughout the entire process.  For label refinement,  we fine-tune a pre-trained ResNet~\cite{resnet} on the ImageNet~\cite{russakovsky2015imagenet} dataset for 80 epochs with a batch size of 256. The predefined confidence thresholds $\lambda_{\text{high}}$ and $\lambda_{\text{low}}$  are determined based on the confidence score that yields the optimal F1-score for synthetic source domain cross-attention maps during detector training. Specifically, we set $\lambda_{\text{high}} = 0.7$ and $\lambda_{\text{low}} = 0.3$  for both YOLOv5 and YOLOv8, while for ViTDet and Faster-RCNN, we assign values of $\lambda_{\text{high}} = 0.95$ and $\lambda_{\text{low}} = 0.5$.

\noindent\textbf{Image Generation:} We employ Stable diffusion V1.4~\cite{rombach2022high}, pre-trained on the
images of size $512 \times 512$, a batch size of 64, and a learning rate of $10^{-6}$ on two RTX A6000 GPUs for approximately 15 epochs. Following section~\ref{subsec:concept localization}, we adopt a two-stage fine-tuning strategy to capture both object and background concepts. For the source domain, the guidance prompt is ``An aerial view image with [$\text{V}_1$] cars in [$\text{V}_2$][id]" for images with small vehicles since most small vehicles fall under the broader classification of cars (see Appendix~\cref{sec:supp-dataset} for more details),  and ``An aerial view image in [$\text{V}_2$][id]" for images without small vehicles, where [id] is replaced with ``New Zealand" or ``DOTA". For the target domain, [$\text{V}_2$] is substituted with token [$\text{V}_3$] to learn unique background concept in target domain, and [id] is replaced with ``Utah". In the first stage, we fine-tune the learnable tokens in the prompt along with the U-Net. In the second stage, we freeze the learned tokens and fine-tune U-Net to better align the model with the data distribution. Each stage is trained for approximately two epochs using a batch size of 8 and a learning rate of $5\times10^{-7}$ on two RTX A6000 GPUs. We synthesize 10\textit{k} images containing cars for both the source and target domain, and 10\textit{k} images without cars for the target domain. Additional implementation details are provided in the Appendix~\cref{sec:supp-implementation}.

\subsection{Comparison with State-of-the-art Methods}

\setlength{\tabcolsep}{2pt}
\begin{table} 
    \centering
    \footnotesize
    \begin{tabularx}{\linewidth}{@{}p{25.6mm}p{21.4mm}>{\centering\arraybackslash}X>{\centering\arraybackslash}X@{}}
        \toprule
        \multirow{2}{*}{\textbf{Method}} & \multirow{2}{*}{\textbf{Backbone}} & \scriptsize{\textbf{LINZ$\rightarrow$UGRC}} & \scriptsize{\textbf{DOTA$\rightarrow$UGRC}} \\
        & & AP$_{50} (\%)$ & AP$_{50} (\%)$ \\
        \midrule
        \midrule

     \multicolumn{4}{@{}l@{}}
    {\textcolor{r2}{\textit{Supervised Object detection (Source-only)}}} \\
        Faster-RCNN~\cite{faster-rcnn} & Faster R-CNN~\cite{faster-rcnn} & 53.1 & 52.5 \\
        YOLOv5~\cite{yolov5} & YOLOv5~\cite{yolov5} & 63.2 & 57.6 \\
        YOLOv8~\cite{yolov5} & YOLOv8~\cite{reis2023real} & 62.9 & 71.4 \\
        ViTDet~\cite{vitdet} & ViTDet~\cite{vitdet} & 55.7 & 50.4 \\
        \midrule
        
        \multicolumn{4}{@{}l@{}}{\textit{Open-set object detection}} \\
        GLIP-T~\cite{li2021grounded} & Swin-T~\cite{Swin-T} & 8.7 & 8.7 \\
        
        OmDet-Turbo~\cite{zhao2024realtimetransformerbasedopenvocabularydetection} & Swin-T~\cite{Swin-T} & 14.4 & 14.4 \\
        OWLV2~\cite{minderer2024scalingopenvocabularyobjectdetection} & CLIP ViT~\cite{CLIP} & 17.9 & 17.9 \\
        \midrule
        \multicolumn{4}{@{}l@{}}{\textit{Unsupervised domain adaptation object detection}}\\
        SIGMA~\cite{li2022sigma} &  Faster R-CNN~\cite{faster-rcnn} & 36.2 & 34.8 \\
        TIA~\cite{TIA} & Faster R-CNN~\cite{faster-rcnn} & 49.2 & 42.4  \\
        Adapt. Teacher~\cite{li2022cross} & Faster R-CNN~\cite{faster-rcnn} & 29.0 &  37.0 \\
        CycleGAN-Turbo~\cite{parmar2024onestepimagetranslationtexttoimage} & YOLOv5~\cite{yolov5} & 56.0 & 60.8  \\
        SSDA-YOLO~\cite{zhou2022ssdayolosemisuperviseddomainadaptive} & YOLOv5~\cite{yolov5} & 52.3 & 49.6  \\
        \midrule
        \multicolumn{4}{@{}l@{}}{\textit{Cross domain weakly supervised object detection}}\\
        OCUD~\cite{ocud} & Faster R-CNN~\cite{faster-rcnn} & 63.1 & 65.3 \\
        H2FA R-CNN~\cite{Xu_2022_CVPR} & Faster R-CNN~\cite{faster-rcnn} & 61.8 & 68.3  \\
        \midrule
        Ours & Faster R-CNN~\cite{faster-rcnn} & \textbf{69.3} & \textbf{75.5} \\
        Ours & YOLOv5~\cite{yolov5} & \textbf{68.8} & \textbf{68.5} \\
        Ours & YOLOv8~\cite{reis2023real} & \textbf{75.4} & \textbf{75.7} \\
        Ours & ViTDet~\cite{vitdet} & \textbf{72.0} & \textbf{67.1} \\
        \bottomrule
    \end{tabularx}
    \caption{\textbf{Cross-domain object detection results}. LINZ to UGRC and DOTA to UGRC. We report the AP$_{50}$  result.}
    \label{tab:cross-domain-sota-satellite}
\end{table}

\Cref{tab:cross-domain-sota-satellite} summarizes the cross-domain detection results for adaptation from the LINZ and DOTA datasets to the UGRC dataset. 
\textcolor{r2}{First, our method consistently outperforms detectors trained solely on source domain data, demonstrating its robustness and generalizability. Second, open-set detectors perform poorly on aerial images, highlighting the lack of training of foundational models in this domain.  
Third, compared to domain adaptation methods~\cite{li2022sigma, TIA, li2022cross, ocud, Xu_2022_CVPR} based on Faster-RCNN~\cite{faster-rcnn}, our pipeline achieves the highest $\text{AP}_{50}$ of 75.4\% for LINZ $\rightarrow$ UGRC, and 75.7\% for DOTA $\rightarrow$ UGRC, surpassing the best unsupervised method by 20.1\% and 33.1\%, and the best weakly supervised method by 6.2\% and 7.2\%, respectively. Similarly,  using YOLOv5~\cite{yolov5} as the backbone, our pipeline outperforms existing methods~\cite{zhou2022ssdayolosemisuperviseddomainadaptive, parmar2024onestepimagetranslationtexttoimage} by 12.8\% and 7.7\%.}  These results strongly corroborate the promise and potential of our method against competing baselines, for the challenging task of vehicle detection from aerial imagery. 

\begin{table}[t]
  \centering
  \footnotesize
  \begin{tabular}{llcc}
  \toprule
    \multirow{2}{*}{\textbf{Method}} & \multirow{2}{*}{\textbf{Backbone}} & \textbf{LINZ$\rightarrow$UGRC} & \textbf{DOTA$\rightarrow$UGRC} \\
    & & AP$_{50} (\%)$ & AP$_{50} (\%)$ \\
    \midrule
    Base & YOLOv5 & 62.1 & 68.0  \\
    Base & YOLOv8 & 69.8 & 74.5  \\
    Stack & YOLOv5 & 68.8 & 68.5 \\
    Stack & YOLOv8 & 75.4 & 75.7 \\
    \bottomrule
  \end{tabular}
\caption{\textbf{Comparison between labeling using single cross-attention map and multi-channel cross-attention maps}. ``Base" denotes using cross-attention maps of word ``car" only. ``Stack" denotes stacked cross-attention maps extracted from the word ``car" and two learnable tokens for the car concept and background.}
\label{tab:abl-stack-heatmap}
\end{table}

\begin{figure*}[t]
    \centering  \includegraphics[width=0.95\textwidth]{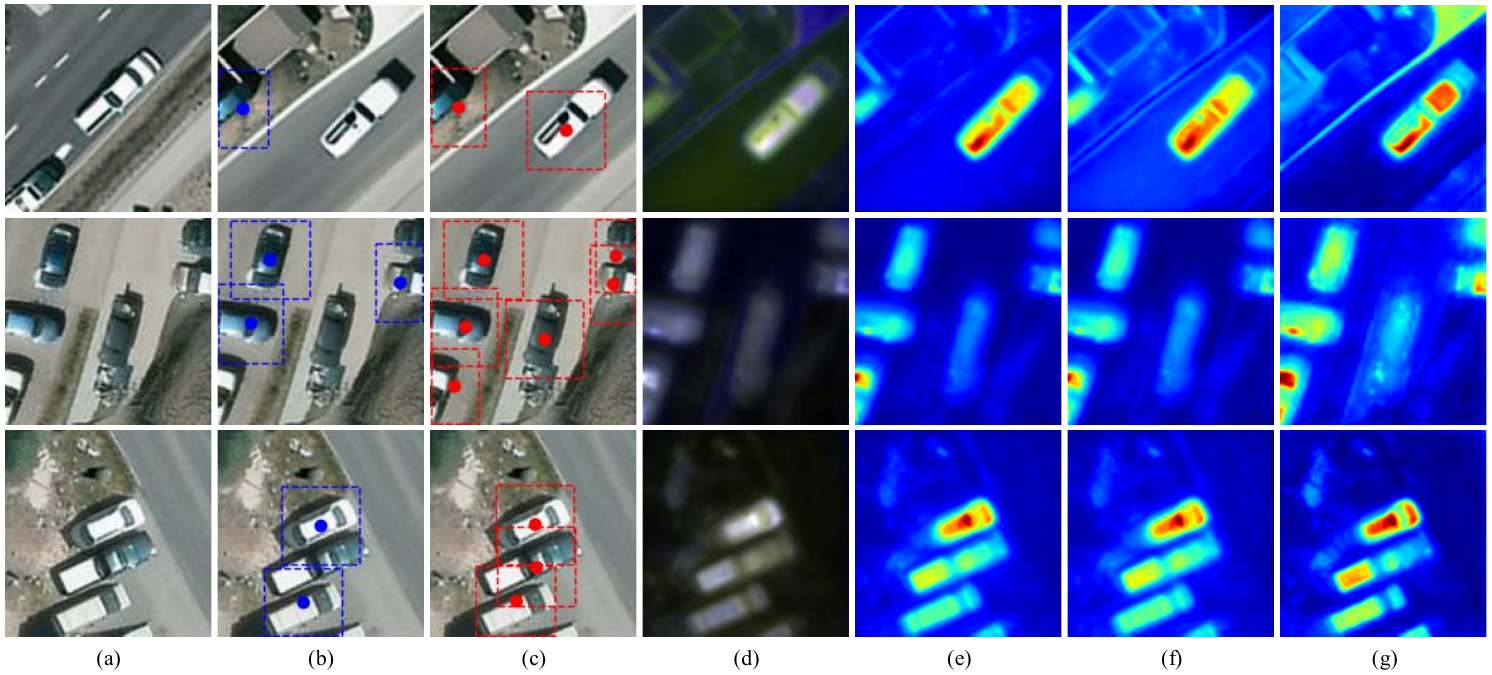}
    \caption{\textbf{Cross-attention maps for different tokens.} We analyze the effectiveness of utilizing multi-channel cross-attention maps by assessing the label quality synthetic UGRC images. (a) Synthetic UGRC images.  (b) Labels generated using only the cross-attention map of the word ``car". (c) Labels generated by using multi-channel heatmaps. (d) Multi-channel cross-attention maps. (e) Cross-attention maps of the word ``car". (f) Cross-attention maps of token [$\text{V}_1$], which is designed to capture the concept of cars. (g) Cross-attention maps of token [$\text{V}_3$] for background concept, which are further inverted for better comparison. In (b) and (c),  bounding boxes with dotted lines denote the predicted pseudo-bounding box labels while dots denote predicted vehicle centers. In (e), (f), and (g), grayscale cross-attention maps are displayed as color heatmaps to highlight the intensity difference.}
\label{fig:ablation-heatmap}
\end{figure*}

\begin{figure}[!tbp]
    \centering  \includegraphics[width=\columnwidth]{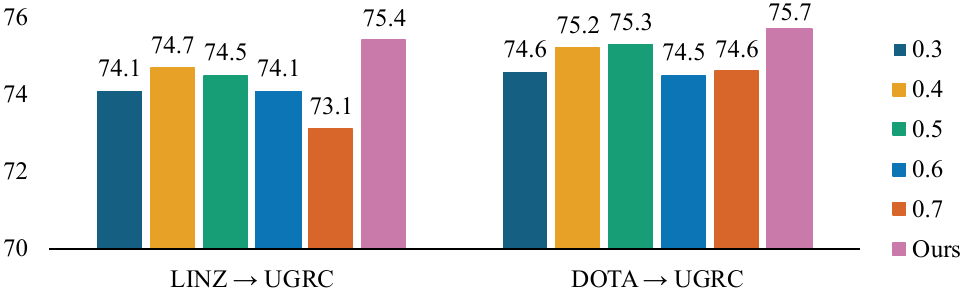}
\caption{
\textbf{Quantitative comparison with varying thresholds.} We report the AP$_{50}$ result.
}
\label{fig:ablation-clf-refine-bar-chart}
\end{figure}

\begin{figure}[!htbp]
    \centering  \includegraphics[width=\columnwidth]{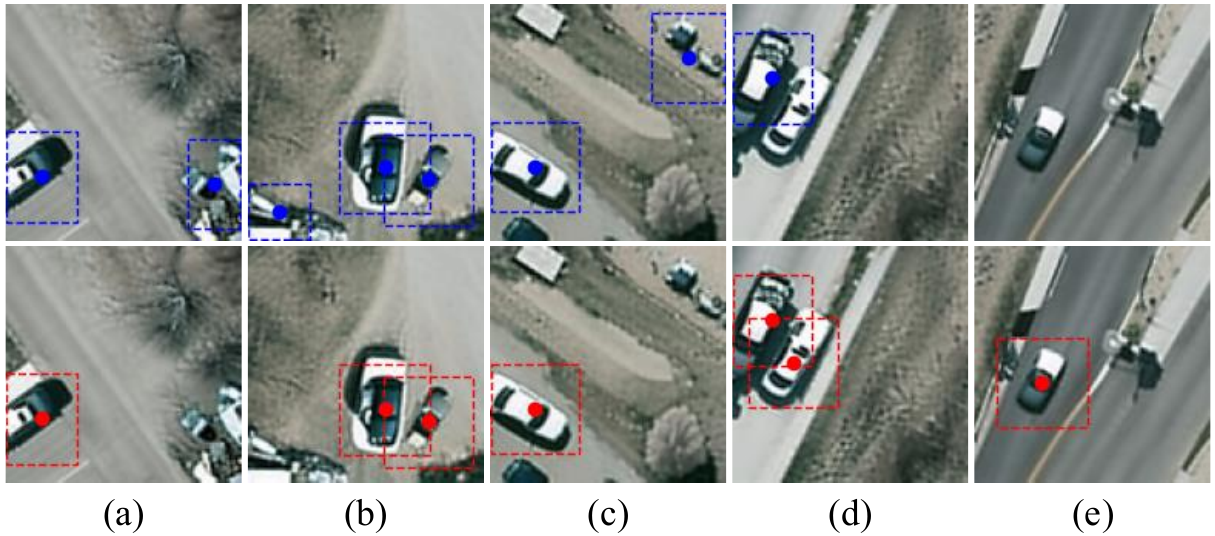}
    \caption{\textbf{Comparison of our method with different thresholds for label quality.} Blue bounding boxes represent pseudo labels generated using fixed thresholds, while red bounding boxes correspond to labels obtained through refinement. The blue and red dots indicate the predicted car centers. (a) to (e) show the results for thresholds of 0.3, 0.4, 0.5, 0.6, and 0.7, respectively.}
\label{fig:ablation-clf-refine-vis}
\end{figure}

\subsection{Ablation Studies}
\subsubsection{Effectiveness of learnable tokens}
We evaluate the effectiveness of stacking three cross-attention maps of the word ``car", learnable tokens of car concept and background for small vehicle detection using YOLOv5 and YOLOv8, as summarized in \Cref{tab:abl-stack-heatmap}. Compared to using only the cross-attention map of the word ``car", incorporating multiple cross-attention maps improves detection accuracy by 5.6\% when adapting from LINZ to UGRC, and by 1.2\% from DOTA to UGRC. Additionally, we provide visualizations of cross-attention maps corresponding to different tokens and analyze their impact on labeling synthetic images, as illustrated in \Cref{fig:ablation-heatmap}. The visualizations reveal that different cross-attention maps tend to focus on distinct regions of the cars. By cross-verifying using multiple maps, the labeling process becomes more robust, leading to improved label quality.

\subsubsection{Effectiveness of label refinement}
To evaluate the effectiveness of fine-tuning a classifier to refine labels, we compare our method against a fixed filtering confidence threshold set at 0.3, 0.4, 0.5, 0.6, and 0.7 using YOLOv8. As shown in \Cref{fig:ablation-clf-refine-bar-chart}, the detection performance varies across different datasets without a consistent pattern about the confidence threshold. When compared to the results obtained using different thresholds, our method demonstrates competitive performance, suggesting that it produces a reliable set of labels.
Additionally,  compared to our method, when setting a lower threshold the detector labels more false positive samples, while setting a higher threshold the model labels fewer cars, as shown in \Cref{fig:ablation-clf-refine-vis}. This indicates our method can be helpful to select a more reliable set of labels.

\section{Conclusions}
\label{sec:conclusions}
This paper introduces a novel approach that leverages diffusion models to automatically generate synthetic aerial view images alongside object
location annotations
based on multi-channel cross-attention maps. Extensive experiments conducted with four
object detectors, as well as comparisons with open-set detectors, unsupervised domain adaptation methods, and weakly supervised models, demonstrate the effectiveness of our method in object detection for aerial imagery. Additionally, we introduce two new aerial view datasets from New Zealand and Utah, including annotations for small vehicle detection tasks. \textcolor{r1}{For future work, we plan to integrate VLLMs trained for aerial view images to automatically identify the presence of vehicles in aerial imagery. This would allow us to generate weak labels within the pipeline itself, enabling a fully unsupervised workflow.}

Despite its strengths, our method has certain limitations that should be considered. First, we sample large aerial view images into smaller patches of resolution $112~\text{px} \times 112~\text{px}$ to mitigate challenges in detecting small objects. This limitation arises because Stable Diffusion progressively downsamples the cross-attention map to a size of $8 \times 8$. Consequently, target objects may occupy less than a $1 \times 1$ grid, making it difficult for the model to generate precise attention maps. Second, our approach may encounter difficulties in handling overlapping objects.  When multiple objects are close together, their cross-attention maps overlap, making it challenging for detectors to separate individual objects when using styleless cross-attention maps for synthetic data annotations. 

\section{Acknowledgement}
\label{sec:acknowledgement}

We thank Jessica Hodgins, Justin Macey, Melanie Danver, Katie Tender, and Eric Yu for their
valuable help in creating the datasets. This work has been
funded by the DEVCOM Army Research Laboratory.
{
    \small
    \bibliographystyle{ieeenat_fullname}
    \bibliography{main}
}

\clearpage
\setcounter{page}{1}
\setcounter{section}{0}


\maketitlesupplementary

\renewcommand{\thesection}{\Alph{section}}

\begin{algorithm}[t]
\caption{Summary of the pipeline}
\textbf{Input:} Source domain data $\mathbf{D}^\mathcal{S}$, target domain data $\mathbf{D}^\mathcal{T}$
\begin{algorithmic}[1]
\State Fine-tune Stable Diffusion~\cite{rombach2022high} on both $\mathbf{D}^\mathcal{S}$ and $\mathbf{D}^\mathcal{T}$ using domain-specific prompts

\State Generate synthetic images $x^\mathcal{GS}$ and extract stacked cross-attention maps $\widetilde{A}^\mathcal{GS}$ for the source domain 

\State Generate synthetic images $x^\mathcal{GT}$ and extract stacked cross-attention maps $\widetilde{A}^\mathcal{GT}$ for the target domain

\State Train a detector ${F}^\mathcal{S}(\cdot; \theta)$ on $\mathbf{D}^\mathcal{S}$

\State Run ${F}^\mathcal{S}(\cdot; \theta)$ on $\mathbf{D}^\mathcal{S}$ to obtain pseudo labels $y^\mathcal{GS}$

\State Train a detector ${F}^\mathcal{A}(\cdot; \theta)$ on $\left(\widetilde{A}^\mathcal{GS}, y^\mathcal{GS}\right)$
\State Run ${F}^\mathcal{A}(\cdot; \theta)$ on $\widetilde{A}^\mathcal{GT}$ to obtain pseudo labels $y^\mathcal{GT}$

\State Train the final detector ${F}^\mathcal{T}(\cdot; \theta)$ on $\left(x^\mathcal{GT}, y^\mathcal{GT}\right)$

\State Test ${F}^\mathcal{T}(\cdot; \theta)$ on real target domain images from $\mathbf{D}^\mathcal{T}$
\end{algorithmic}
\label{alg:supp-method}
\end{algorithm}

\section{Methods}
\label{sec:supp-method}
\textcolor{r3}{In this section, we present a summary of steps regarding our proposed pipeline, as shown in \Cref{alg:supp-method}.}

\begin{figure}[t]
    \centering  \includegraphics[width=0.76\columnwidth]{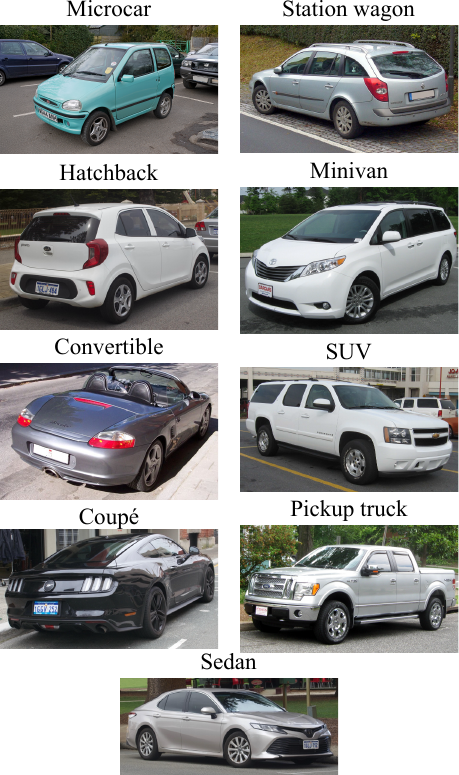}
    \caption{Vehicles belonging to the object class \textit{small vehicle}.
    }
    \label{fig:supp-Small-vehicle}
\end{figure}


    
    

\begin{figure*}[htbp]
    \centering
    \makebox[\textwidth]{%
        \begin{subfigure}[t]{0.5\textwidth}
            \centering    \includegraphics[height=11cm,valign=t]{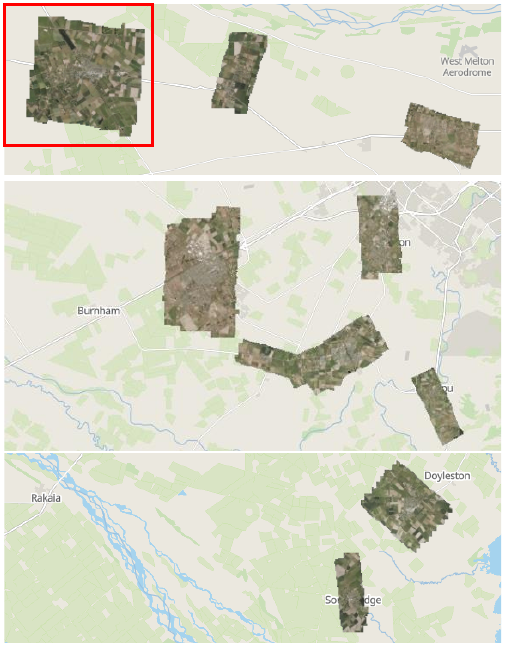}
            \caption{Selwyn (New Zealand)}  
        \end{subfigure}
        
        \begin{subfigure}[t]{0.5\textwidth}
            \centering   \includegraphics[height=11cm,valign=t]{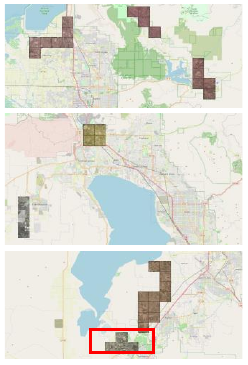}
            \caption{Utah (USA)}
        \end{subfigure}
    }
    \caption{Geographic regions where we construct LINZ and UGRC datasets. Red bounding boxes denote the testing area.}
    \label{fig:supp-linz-utah-train-test-split}
\end{figure*}

\begin{figure*}[t]
    \centering  \includegraphics[width=\textwidth]{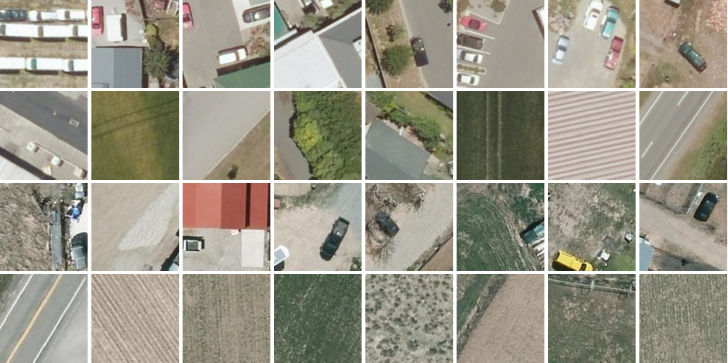}
    \caption{
        \textbf{Examples of images from our real-world datasets.}
        (\textit{first row})~LINZ images containing small vehicles;
        (\textit{second row}) LINZ images without vehicles;
        (\textit{third row})~UGRC images containing small vehicles;
        (\textit{fourth row}) UGRC images without vehicles;
    }
    \label{fig:LINZ-UGRC_MultipleExamples}
\end{figure*}

\begin{figure*}[!htbp]
    \centering  \includegraphics[width=\textwidth]{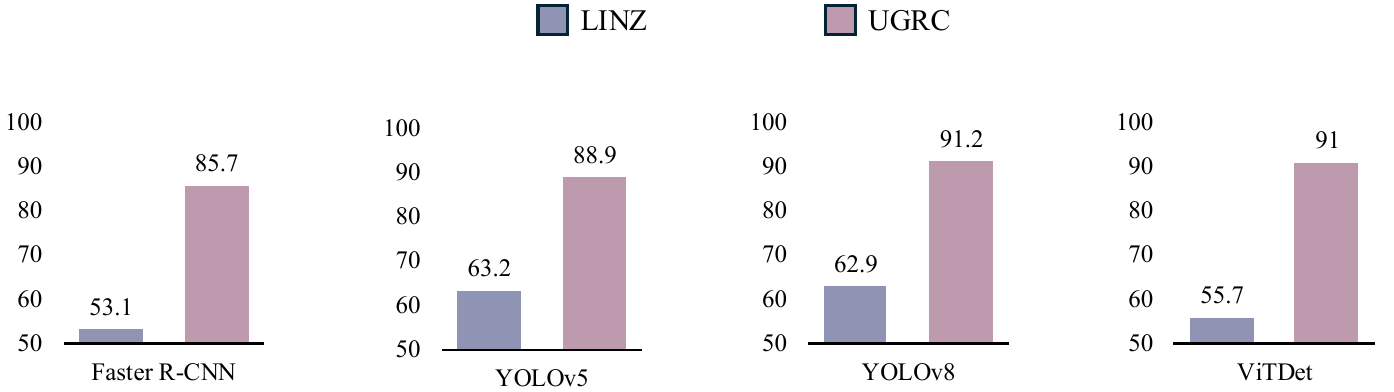}
    \caption{
    \textbf{Comparison between cross-dataset generalization and within-dataset performance.}
    The purple bars represent the model trained on the LINZ dataset and evaluated on the UGRC dataset, while the pink bars correspond to both training and testing conducted on the UGRC dataset. We report the $\text{AP}_{50}$ result.
    }
    \label{fig:supp-compare-baseline}
\end{figure*}

\section{Datasets}
\label{sec:supp-dataset}


In this section, we provide additional details and examples of the LINZ and UGRC datasets. \Cref{fig:supp-linz-utah-train-test-split} illustrates the geographic regions from which our data samples were obtained. The LINZ online platform captured aerial imagery from nine areas in Selwyn. For dataset construction, we designated one of these nine areas as the testing region, from which test set images were sampled, while the remaining eight areas were used for training and validation samples. Similarly, the UGRC online platform collected aerial imagery from seven regions in Utah. One of these seven areas was designated as the testing region, with the remaining six serving as sources for training and validation data. This spatial partitioning strategy ensures that our datasets do not suffer from data leakage, as the training and testing areas are spatially independent. Within each area, we randomly sample square images of size 112~px $\times$ 112~px. Due to the sampling strategy, a single vehicle can appear in multiple images. \Cref{fig:supp-Small-vehicle} presents the subcategories within the \textit{small vehicle} class. Except for the \textit{Pickup truck} category, all other small vehicle subcategories fall under the broader classification of cars.  Accordingly, we utilize the word ``car" in prompts to guide the Stable Diffusion~\cite{rombach2022high} during image generation, leveraging its pre-trained perceptual understanding related to cars. \Cref{fig:LINZ-UGRC_MultipleExamples} provides visual examples of both LINZ and UGRC images, \textcolor{r1}{highlighting distinct visual characteristics: UGRC includes a notable proportion of off-road vehicles, reflecting its sandy and rocky terrain, while LINZ images primarily feature urban vehicles within structured road networks. Lastly, \Cref{fig:supp-compare-baseline} compares the performance of four object detectors under two settings: cross-dataset evaluation (trained on LINZ and tested on UGRC) and within-dataset evaluation (trained and tested on UGRC). The results show that within-dataset performance surpasses cross-dataset performance by at least 25.7\% higher $\text{AP}_{50}$ across all detectors, underscoring a significant domain gap between the two datasets.} 

\begin{figure}%
    \centering
    \subfloat[LINZ - BLIP2 Image Captioning]{\includegraphics[width=\columnwidth]{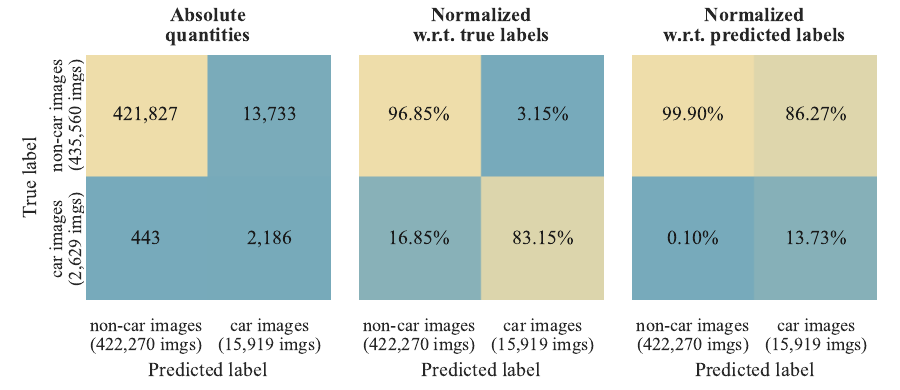}}%
    
    \subfloat[LINZ - Kosmos2 Image Captioning]{\includegraphics[width=\columnwidth]{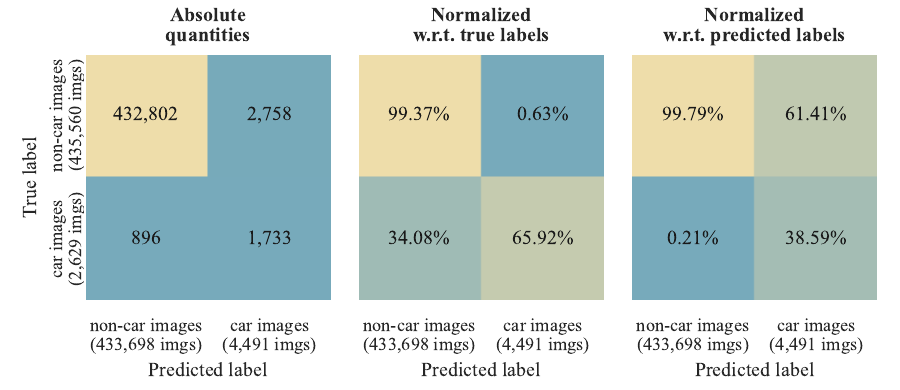}}%

    \subfloat[UGRC - BLIP2 Image Captioning]{\includegraphics[width=\columnwidth]{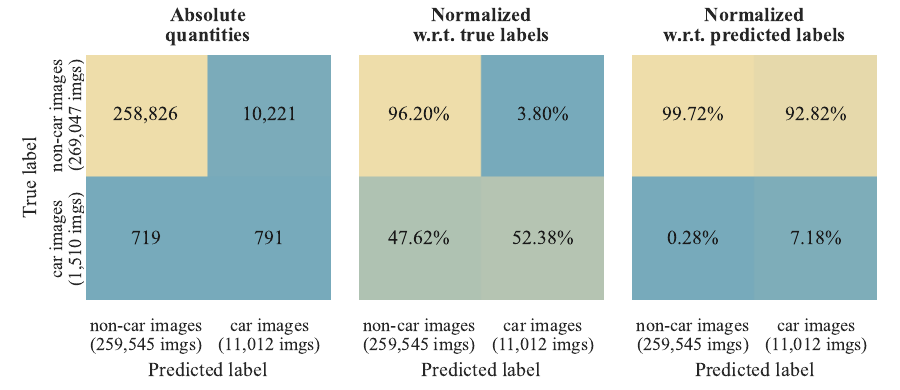}}%
    
    \subfloat[UGRC - Kosmos2 Image Captioning]{\includegraphics[width=\columnwidth]{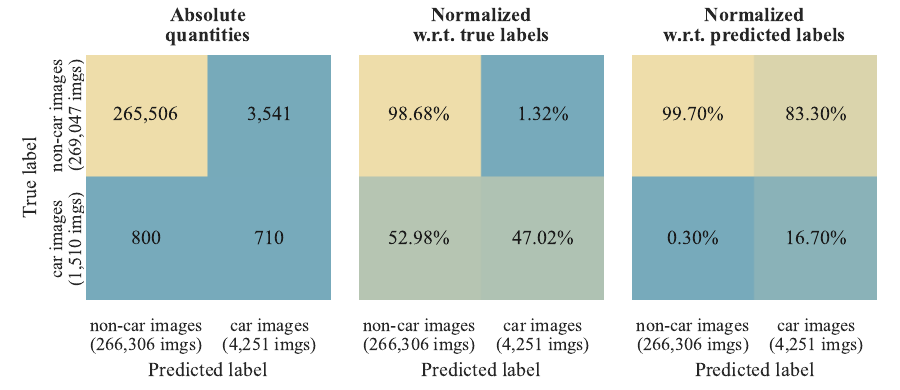}}%
    
    \caption{
        VLMs captioning capabilities analysis tested on LINZ and UGRC datasets.
    }%
    \label{fig:supp-VLM_clf}%
\end{figure}

\begin{figure}%
    \centering
    \includegraphics[width=\columnwidth]{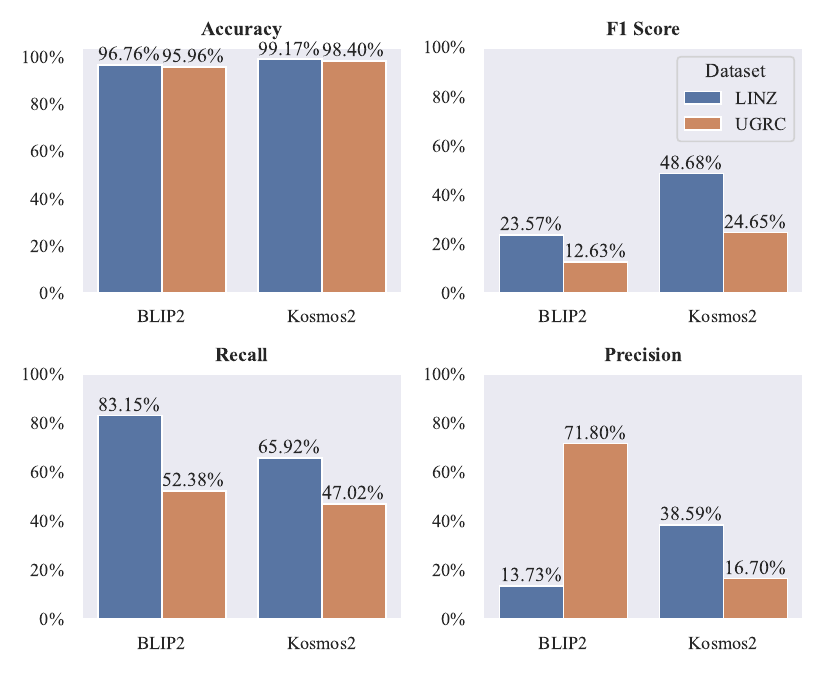}
    
    \caption{Two popular VLLMs (BLIP2 and Kosmos2) tested as zero-shot image car presence classifier. The severe imbalance of the positive and negative classes causes the high levels of accuracy, which is deceptive. F1 score, Precision and Recall metrics clearly show that the classification quality is less than ideal.}%
    \label{fig:supp-VLM_metrics}%
\end{figure}

\setlength{\tabcolsep}{2pt}
\begin{table}[htbp]
    \centering
    \footnotesize
    \begin{tabular}
    {lccc}
        \toprule
        \multirow{2}{*}{\textbf{Method}} & \multirow{2}{*}{\textbf{Vision Backbone}} & \multicolumn{2}{c}{\scriptsize{\textbf{LINZ$\rightarrow$UGRC}}} \\
        & & Precision(\%) & Recall(\%) \\
        \midrule
        \midrule
        \multicolumn{4}{l}{\textit{Vision Large Language Model}} \\
        Gemini 1.5 Flash~\cite{gemini1.5} & - & 2.9 & 44.5 \\
        Gemini 2.0 Flash-Lite~\cite{gemini2} & - & 6.6 & 26.3  \\
        InternVL3-8B~\cite{internvl3} & InternViT~\cite{internvl3} & 4.7  & 22.0 \\
        Qwen2.5-VL-7B~\cite{qwen25vl} & ViT~\cite{qwen25vl} & 0.4 & 4.8 \\        
        DeepSeek-VL2-Tiny~\cite{deepseekvl2} & SigLIP-SO400M~\cite{siglip} & 9.2  & 26.8 \\
        LLaVA-NeXT~\cite{llavanext} & CLIP ViT~\cite{CLIP} & 5.5 & 4.7 \\
        \midrule
        Ours & Faster R-CNN~\cite{faster-rcnn} & \textbf{63.8} & \textbf{68.2} \\
        Ours & YOLOv5~\cite{yolov5} & \textbf{67.2} & \textbf{67.3} \\
        Ours & YOLOv8~\cite{reis2023real} & \textbf{70.0} & \textbf{76.3} \\
        Ours & ViTDet~\cite{vitdet} & \textbf{72.0} & \textbf{67.1} \\
        \bottomrule
    \end{tabular}
    \caption{Comparison between our methods and VLLMs on UGRC dataset. We report the precision and recall metrics.}
    \label{tab:supp-cross-domain-vllm-satellite}
\end{table}

\section{Limitation of Foundation Models}
\label{sec:supp-failure_case}
 In this section, we provide more examples of limitations of various types of foundation models, including open-set detectors, diffusion models, and vision large language models, when detecting small vehicles in aerial view images.

\subsection{Open-set Detectors}
\begin{figure*}[t]
    \centering  \includegraphics[width=\textwidth]{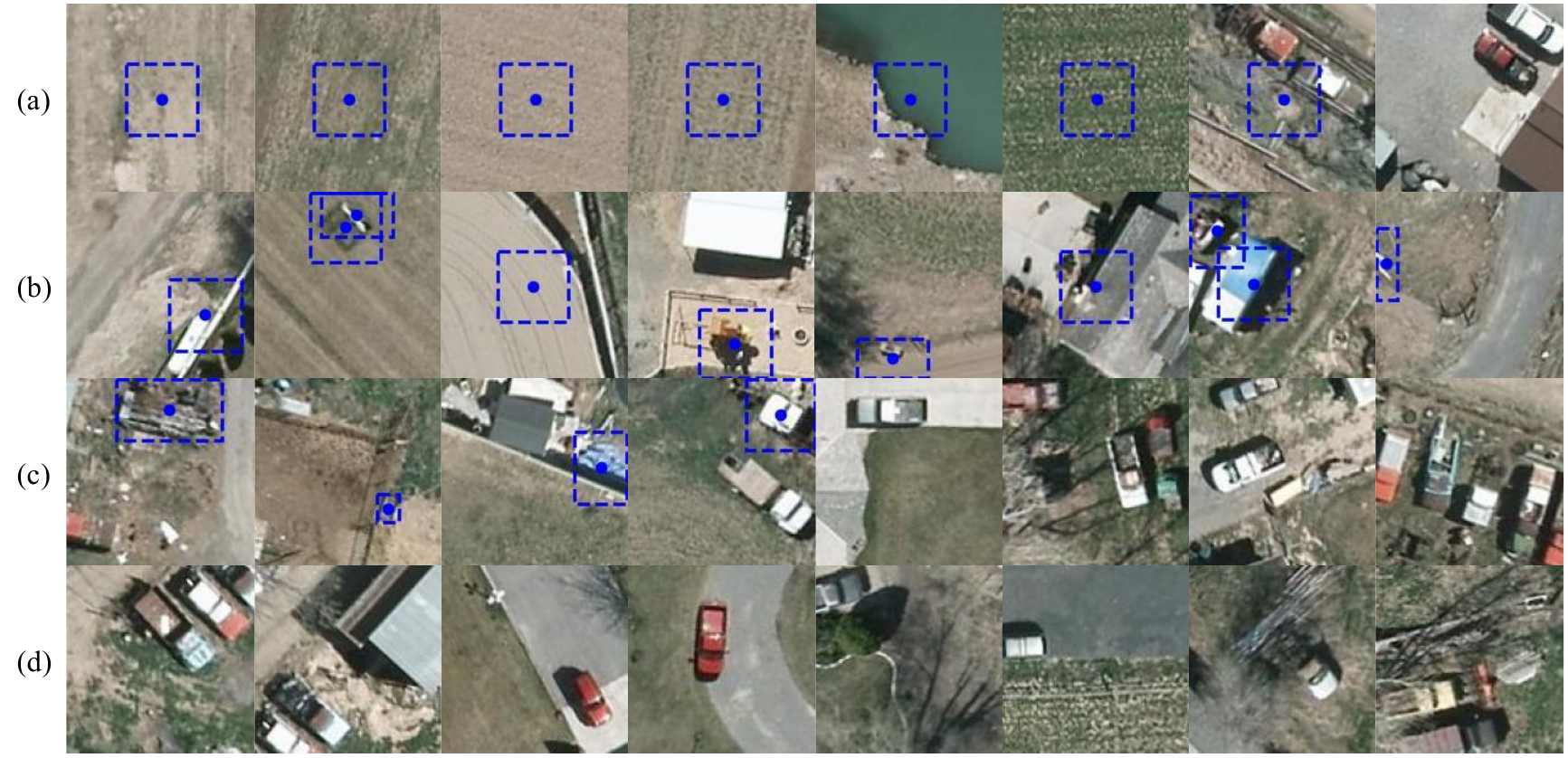}
    \caption{Failure cases of Open-set detectors. (a) Detection results of Grounding-DINO. (b) Detection results of Omdet-Turbo. (c) Detection results of OWLV2. (d) Detection results of Owlvit. The blue bounding boxes with dotted lines denote the predicted pseudo bounding box labels while the dots denote the predicted car centers.}
\label{fig:failure-open-set-detector}
\end{figure*}

As shown in Figure~\ref{fig:failure-open-set-detector}, Grounding-DINO~\cite{liu2024groundingdinomarryingdino} often detects cars in background images. OmDet-Turbo~\cite{zhao2024realtimetransformerbasedopenvocabularydetection} and OWLv2~\cite{minderer2024scalingopenvocabularyobjectdetection} often produce false positives by misclassifying objects such as rectangular tanks and boxes, which share visual similarities with cars. OWL-ViT~\cite{minderer2022simpleopenvocabularyobjectdetection} fails to detect any cars in aerial images, highlighting its limitations in this specific context. These findings underscore the challenges faced by open-set object detectors in accurately identifying vehicles in aerial imagery.

\subsection{Diffusion Models}
\begin{figure*}[t]
    \centering  \includegraphics[width=\textwidth]{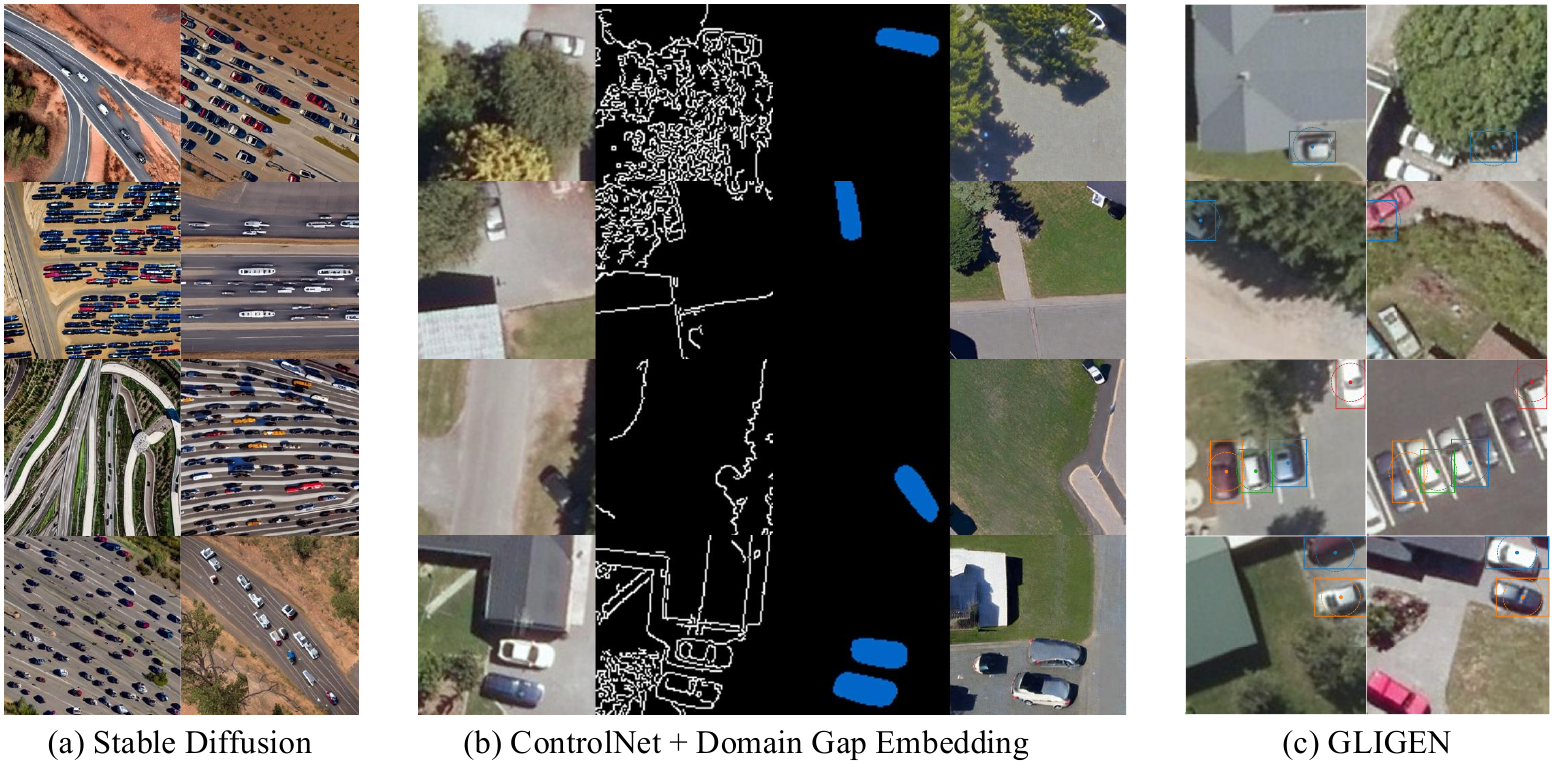}
    \caption{Failure cases of diffusion models. (a) Images generated by pre-trained Stable Diffusion V1.4. (b) Images generated by pre-trained Stable UnCLIP~\cite{ramesh2022hierarchicaltextconditionalimagegeneration} with canny edge maps, semantic segmentation maps, and average CLIP image embedding difference between the LINZ and the UGRC dataset as conditions. From left to right: Real images from LINZ dataset, edge maps, semantic segmentation maps, and synthetic images.  (c) Images generated by GLIGEN. Left: Real images from the LINZ dataset. Right: Synthetic UGRC images. The bounding boxes that have the same color in left and right images correspond to the same location. }
\label{fig:failure-generative-model}
\end{figure*}

As shown in Figure~\ref{fig:failure-generative-model} (a), when employing the pre-trained Stable Diffusion~\cite{rombach2022high} with the prompt ``an aerial image with cars in Utah", the model fails to understand the geographical reference to ``Utah" and does not generate images reflective of the state's landscape. Additionally, Stable Diffusion struggles with generating small objects such as cars in aerial images, resulting in low-quality vehicle depictions. ControlNet~\cite{ControlNet} may also fail to effectively guide the generation direction, leading to outputs that do not align with the provided edge and segmentation map conditions. As shown in Figure~\ref{fig:failure-generative-model} (b), DoGE~\cite{wang2024domain} frequently fails to generate cars and surroundings in the precise locations corresponding to the input LINZ image under the guidance of canny edges and semantic segmentation maps. Moreover, without finetuning on UGRC images, DoGE fails to encode the domain difference by modeling the average CLIP~\cite{CLIP} image embedding difference and is unable to produce images that accurately reflect Utah’s landscape. Figure~\ref{fig:failure-generative-model}(c) illustrates that even after fine-tuning GLIGEN~\cite{GLIGEN} on UGRC data, the model frequently fails to comply with the specified bounding box layout conditions. It often generates extraneous cars outside the designated bounding boxes or omits cars within the expected bounding areas. These limitations highlight the challenges associated with ensuring diffusion models faithfully adhere to spatial and semantic constraints in conditioned image generation.

\subsection{Vision Large Language Models}

\textcolor{r3}{ We evaluate the capabilities of Vision Large Language Models (VLLMs) for car presence classification and center localization (VLLMs), as shown in \Cref{fig:supp-VLM_clf} and \Cref{tab:supp-cross-domain-vllm-satellite}. For classification, } 
BLIP2~\cite{li2023blip2} generates captions based on the images while Kosmos2~\cite{Kosmos2} is prompted to complete the sentence ``an aerial image of \{\}".
We consider a model to have predicted the presence of a car in an image if the generated caption includes the word ``car". As shown in Figure~\ref{fig:supp-VLM_clf}~(a), only 13.73\% of the images predicted to contain cars by BLIP2 are true positives in the LINZ dataset. A similar trend is observed in Figure~\ref{fig:supp-VLM_clf} (b), (c) and (d), where the true positive rates for predicted car images are 38.59\%, 7.18\%, and 16.70\%, respectively. Furthermore, among all images that contain cars, only 52.38\% are correctly identified by BLIP2 in the UGRC dataset, as shown in Figure~\ref{fig:supp-VLM_clf} (c). A similar trend is observed in Figure~\ref{fig:supp-VLM_clf} (d), where 47.02\% of all images that actually contain cars are correctly predicted by Kosmos2. Figure~\ref{fig:supp-VLM_metrics} supports that VLLMs struggle to accurately detect cars in aerial imagery by presenting the F1 score, Precision, and Recall metrics of their classification performance. 

\textcolor{r3}{For localization, we assess the detection performance of VLLMs as shown in \Cref{tab:supp-cross-domain-vllm-satellite}. Since VLLMs do not provide confidence scores for the predicted bounding boxes, we define \textit{detection accuracy} as the proportion of predicted bounding boxes that achieve an Intersection over Union (IoU) greater than 0.5 with at least one ground truth bounding box. Based on this definition, we compute precision and recall, which are reported in \Cref{tab:supp-cross-domain-vllm-satellite}. To establish pseudo labels on the UGRC test set for our method, we set the detection threshold according to the highest F1 score achieved by each detector on UGRC test set. Under this setting, VLLMs exhibit significantly lower performance compared to our method, often producing a large number of false positives in aerial imagery.  These findings highlight the current limitations of pre-trained VLLMs in accurately detecting and localizing vehicles in aerial imagery.} 

\begin{figure*}[t]
    \centering  \includegraphics[width=\textwidth]{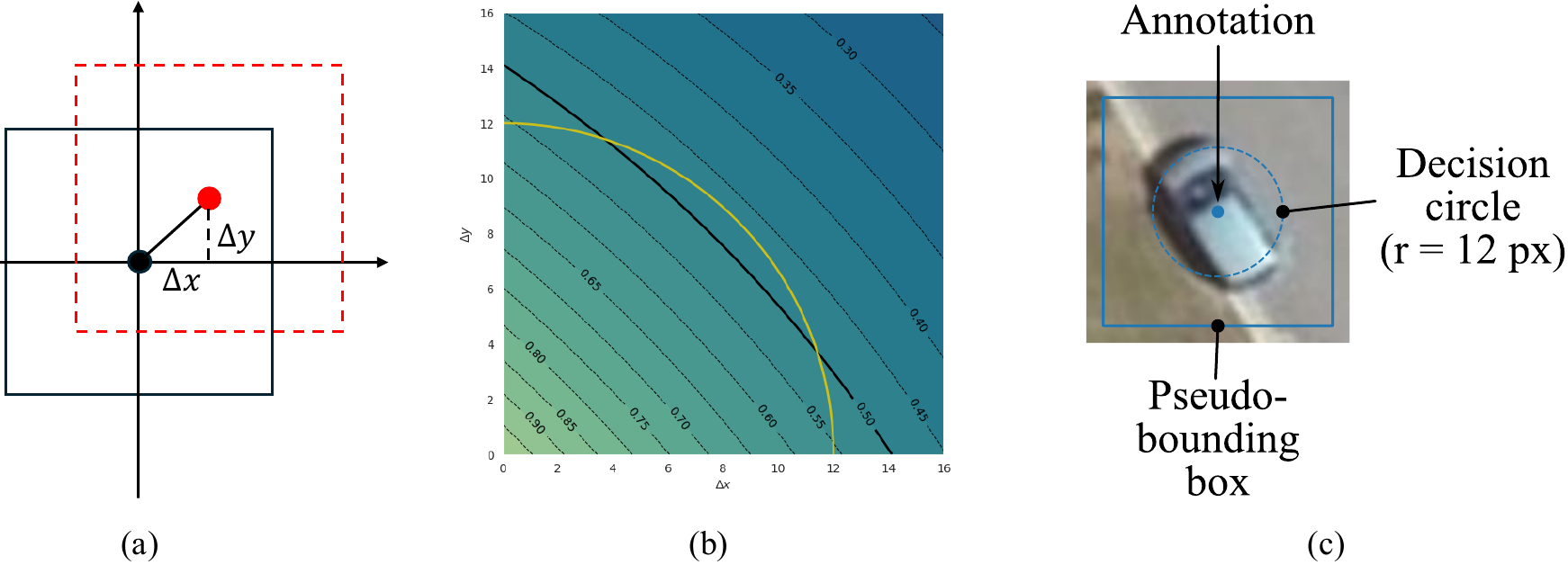}
    \caption{Illustration of how we obtain the 42.36~px bounding box size. (a) The black bounding boxes denote the ground truth pseudo bounding box labels while the red bounding boxes denote the predicted  pseudo bounding box labels. The dots denote the corresponding  centers. $\Delta x$ and $\Delta y$ denote the x-and y-coordinate difference between the ground truth center and the predicted center.  (b) Isocontour of Intersection of Union (IoU). The yellow arc is $\frac{1}{4}$ of the decision circle with a 12~px radius while the black curve represents the isocontour where $\text{IoU}=0.5$. (c) An example of a decision circle with a radius of 12~px centered at the car's center with the corresponding 42.36~px pseudo-bounding box. }
\label{fig:supp-bbox}
\end{figure*}

\setlength{\tabcolsep}{2pt}
\begin{table}[t]
\centering
\footnotesize
\begin{tabular}{c|c|cc|cc|cc}
\hline
\multicolumn{1}{c|}{\multirow{2}{*}{Backbone}} & \multirow{2}{*}{Task}   & \multicolumn{2}{c|}{Stage 1} & \multicolumn{2}{c|}{Stage 2} & \multicolumn{2}{c}{Stage 3} \\
\multicolumn{1}{c|}{}         &   &  bs & lr & bs & lr & bs & lr \\ \hline

Faster-RCNN & LINZ $\rightarrow$ UGRC & 1024 &  0.2  & 192 & 0.02  & 512 & 0.2 \\

Faster-RCNN & DOTA $\rightarrow$ UGRC & 1024 & 0.2  & 192  & 0.001  & 1024 & 0.2 \\

YOLOv5 & LINZ $\rightarrow$ UGRC & 1600 &  0.001  & 192 & 0.0001  & 2400 & 0.001 \\

YOLOv5 & DOTA $\rightarrow$ UGRC &  2400 &  0.001 & 192  & 0.0001  & 2400 & 0.001 \\

YOLOv8 & LINZ $\rightarrow$ UGRC &  4096 & 0.001 &  192  & 0.0001  & 1024 & 0.001  \\

YOLOv8 & DOTA $\rightarrow$ UGRC & 4096 & 0.001 &  192  & 0.0001  & 2048 & 0.001  
\\

ViTDet & LINZ $\rightarrow$ UGRC & 192 & 0.001 &  96  & 0.0001  & 192 & 0.001  \\

ViTDet & DOTA $\rightarrow$ UGRC & 192 & 0.0001 &  96  & 0.001  & 192 & 0.0001
\\
\hline
\end{tabular}
\caption{Training parameters of each stage, where ``bs" denotes the batch size and ``lr" denotes the base learning rate, which will be scaled during training based on batch size following the MMDetection framework.}
\label{tab:supp-training-param}
\end{table}

\section{More Implementation Details}
\label{sec:supp-implementation}

\subsection{Decision circle and Pseudo-bounding box label}
In this section, we provide a detailed explanation for defining the pseudo-bounding box size to be 42.36~px. As illustrated in Figure~\ref{fig:supp-bbox} (b), any point $p$ within the region enclosed by the isocontour of $\text{IoU} = \alpha$ represents a predicted pseudo-bounding box centered at $p$ with an $\text{IoU} \geq \alpha$ relative to the ground truth pseudo-bounding box. Ideally, all true positive predicted centers should be contained within the decision circle.  However, no isocontour perfectly fits the arc of the decision circle.
To minimize this discrepancy, we determine the square pseudo-bounding box size $a$ such that the area enclosed by the isocontour at $\text{IoU} = 0.5$ matches $\frac{1}{4}$ of the decision circle. Let $\Delta x$ and $\Delta y$ denote the x- and y-coordinate differences, respectively, between the centers of the ground truth and predicted pseudo-bounding boxes, as shown in Figure~\ref{fig:supp-bbox} (a). Without loss of generality, we assume the predicted pseudo-bounding box center lies in the first quadrant. The IoU can then represented as $\text{IoU} = \frac{(a-\Delta x)(a-\Delta y)}{2a^2 - (a-\Delta x)(a-\Delta y)}$. By setting $\text{IoU}=0.5$, we solve for $\Delta y$ in terms of $\Delta x$, treating $a$ as a constant, which can be represented as $\Delta y = \frac{a(a-3\Delta x)}{3(a-\Delta x)}$. We then integrate $\Delta y$ with respect to $\Delta x$ to compute the area under the isocontour of $\text{IoU}=0.5$, which is a function of $a$. Finally we equate this integral to $\frac{1}{4}$ of the area of the decision circle and solve for $a$.
\subsection{Multi-Stage Training}
In this section, we provide more details regarding the training process of the detectors. As outlined in \cref{subsec:automatic labeling}, the labeling of synthetic target domain (UGRC) images is conducted in three stages. In the first stage, we train a detector $F^S$ on fully annotated real source domain data (LINZ or DOTA) and subsequently generate pseudo labels for the synthetic source domain images. In the second stage, we train another detector $F^A$ on the multi-channel cross-attention maps of synthetic source domain images and use it to predict pseudo labels for the multi-channel cross-attention maps of synthetic target domain images. Finally, in the third stage, we train a detector $F^T$ on the synthetic target domain images. For Faster-RCNN~\cite{faster-rcnn}, we use ResNet50~\cite{resnet} as backbone. For YOLOv5~\cite{yolov5}, we utilize the YOLOv5-M variant, while for YOLOv8~\cite{reis2023real}, we employ the YOLOv8-M variant. For ViTDet~\cite{vitdet}, we disable the mask head. In all training stages, we scale the image resolution to $128~\text{px} \times 128~\text{px}$, as YOLOv5 requires input dimensions to be multiples of 32. The specific training parameters for each stage are provided in Table~\ref{tab:supp-training-param}. Except for these adjustments, we adhere to the MMDetection~\cite{mmdetection} framework for implementation.

\begin{table}[!t]
    \centering
    \setlength{\tabcolsep}{4pt}{
    \begin{tabular}{cccc}
        \toprule
    Backbone & $A_\text{c}+A_\text{fg}$ & $A_\text{c}+A_\text{bg}$ & $A_\text{c}+A_\text{fg}+A_\text{bg}$ \\
        \midrule
        YOLOv5~\cite{yolov5} & 63.7 & 65.5 & \textbf{68.8} \\
        YOLOv8~\cite{reis2023real} &  69.1 & 73.1 & \textbf{75.4} \\
    \bottomrule
    \end{tabular}}
\caption{Comparison of different cross-attention map configurations for adaptation from LINZ to UGRC.}
\label{tab:supp-abl-stack-heatmap}
\end{table}

\section{More Ablation Studies}
\textcolor{r2}{In this section, we present two additional ablation studies to further validate the effectiveness of our proposed pipeline. First, we investigate the impact of stacking different combinations of cross-attention maps. Specifically, we compare our approach with alternative configurations that stack two channels of the object category cross-attention map $A_{\text{c}}$ and one channel from either the learned foreground cross-attention map
$A_{\text{fg}}$ or background cross-attention map $A_{\text{bg}}$, ensuring compatibility with object detectors that accept only three-channel inputs. As shown in \Cref{tab:supp-abl-stack-heatmap}, integrating both background and foreground information, as in our method, yields the best performance in $\text{AP}_{50}$.} 

\textcolor{r2}{Second, we analyze the effectiveness of our two-stage design, which first fine-tunes Stable Diffusion~\cite{rombach2022high} on both source and target domain datasets, and then introduces learnable tokens to extract cross-attention maps that capture both foreground and background information. We compare this design with a one-stage baseline that jointly fine-tunes Stable Diffusion and learns tokens simultaneously. The two-stage setup is motivated by the limited localization ability of unseen prompts in pre-trained Stable Diffusion models when applied to aerial view images. For instance, prompts such as ``an aerial view image with cars in Utah" often fail to localize vehicles, leading to inaccurate attention and suboptimal token learning. By first fine-tuning the model to better align the concept of ``cars" with actual vehicle locations, we enable subsequent token learning to more precisely focus on relevant regions. Experimental results show that the one-stage pipeline yields an 8.5\% lower $\text{AP}_{50}$ on YOLOv5~\cite{yolov5} when adapting from LINZ to UGRC.
}

\end{document}